\documentclass[runningheads]{llncs}

\usepackage{eccv}
\usepackage{eccvabbrv}
\usepackage{graphicx}
\usepackage{booktabs}
\usepackage{multirow}
\usepackage{amsmath}
\usepackage{amssymb}
\usepackage{algorithm}
\usepackage{algorithmic}
\usepackage{hyperref}

\hypersetup{
  colorlinks=true,
  linkcolor=blue,
  citecolor=blue,
  urlcolor=blue,
  pdfauthor={
    Divine Yao Agbobli;
    Geoffery Eyram Agorku;
    Israel Afriyie;
    Kwadwo Amankwah-Nkyi;
    Marvin Osei-Kuffour;
    Richmond Owusu Duah;
    Bright Seglah;
    Kelvin Asamoah Terkper;
    Kwabena Amoako Adjei
  },
  pdftitle={
    DRAFE: Domain-Robust Asymmetric Fusion of Heterogeneous Detection
    Transformers for Cross-City Fine-Grained Traffic Object Detection
  }
}

\newcommand{\LWXA}{LW-X\textsubscript{A}}
\newcommand{\LWXB}{LW-X\textsubscript{B}}
\newcommand{\RFB}{RF-B}

\begin{document}
\sloppy

\title{
  DRAFE: Domain-Robust Asymmetric Fusion of Heterogeneous Detection
  Transformers for Cross-City Fine-Grained Traffic Object Detection
}
\titlerunning{DRAFE for Cross-City Fine-Grained Object Detection}

\author{%
  Divine Yao Agbobli\inst{1} \and
  Geoffery Eyram Agorku\inst{2} \and
  Israel Afriyie\inst{3} \and
  Kwadwo Amankwah-Nkyi\inst{4} \and
  Marvin Osei-Kuffour\inst{5} \and
  Richmond Owusu Duah\inst{5} \and
  Bright Seglah\inst{4} \and
  Kelvin Asamoah Terkper\inst{6} \and
  Kwabena Amoako Adjei\inst{5}%
}
\authorrunning{Agbobli et al.}

\institute{%
  Iowa State University, Ames, IA, USA \and
  University of Arkansas, Fayetteville, AR, USA \and
  Parsons Corporation, USA \and
  Jacobs, USA \and
  University of South Florida, Tampa, FL, USA \and
  Neel-Schaffer, Inc., USA \\
  \email{
    dya@iastate.edu,
    gagorku@uark.edu,
    Israel.afriyie@parsons.com,
    kwadwo.amankwahnkyi@jacobs.com,
    moseikuffour@usf.edu,
    rowusuduah@usf.edu,
    Bright.Seglah@jacobs.com,
    kelvin.terkper@neel-schaffer.com,
    Kwabenaadjei@usf.edu
  }%
}

\maketitle

\begin{abstract}
Deep learning-based object detectors are fundamental to intelligent
transportation systems, enabling traffic monitoring, vehicle analytics,
and infrastructure management. However, achieving both fine-grained
vehicle recognition and robust cross-city domain generalization remains
challenging. We present the Domain-Robust Asymmetric Fusion Ensemble
(DRAFE), which combines independently trained LW-DETR and RF-DETR
detectors for cross-city fine-grained traffic object detection. DRAFE
employs a two-stage training strategy that first pretrains complementary
detectors on diverse public traffic datasets using pseudo-label expansion
and human-in-the-loop annotation refinement, producing a curated corpus
of 6,049 images and 203,619 annotations, before challenge-compliant
fine-tuning on the Project Hafnia Track~6 dataset. At inference, DRAFE
applies anchor-conditioned class-consistent matching, reliability-weighted
coordinate fusion, agreement-aware confidence recalibration, and
complementary hypothesis recovery. On AI City Challenge 2026 Track~6,
DRAFE achieves 0.4022 mAP, ranks sixth among 25 participating teams, and
improves by 0.0553 mAP over a preliminary ensemble evaluated under
identical benchmark conditions. Code is available in the
\href{https://github.com/dyagbobli/VisionOps-Trainer}{VisionOps Trainer
repository}.

\keywords{
  Fine-grained object detection \and
  Domain generalization \and
  Ensemble detector \and
  Intelligent transportation systems \and
  Privacy-preserving learning
}
\end{abstract}

\section{Introduction}
\label{sec:intro}

Object detection and classification are fundamental components of
intelligent transportation systems (ITS), enabling traffic-state
estimation, incident detection, infrastructure monitoring, and autonomous
traffic management
\cite{milestone2026training,robinson2026rfdetr}.
Recent advances through convolutional neural networks,
transformer-based architectures, and large-scale pretraining have
substantially improved detection performance
\cite{lin2017focal,najeeb2022finegrained,peng2025dfine,
solovyev2021wbf,tang2025aicity9,vidit2023clip}.

Despite these advances, deploying detectors beyond their training
environments remains challenging. Variations in camera viewpoints,
roadway geometry, traffic density, and regional vehicle distributions
introduce domain shifts that can degrade performance in previously unseen
environments \cite{chen2018domain,ren2015faster}.

Track~6 of the 10th AI City Challenge provides a practical benchmark for
this problem \cite{aicity2026track6}. Unlike coarse detection tasks,
Track~6 requires fine-grained recognition across ten visually similar
classes: Car, Pickup Truck, Single Truck, Combination Truck, Trailer,
Heavy-Duty Vehicle, Van, Motorcycle, Bicycle, and Person. The dataset
exhibits a pronounced long-tailed distribution, leaving several categories
underrepresented
\cite{ghiasi2021copypaste,hao2024simplifying,hosang2017learning}.
Furthermore, Track~6 uses Project Hafnia, a GDPR-compliant
Training-as-a-Service platform that prevents participants from downloading
or directly accessing the underlying challenge imagery
\cite{lavoie2025large,milestone2024dataset,milestone2025hafnia}. Models
are trained within Hafnia on labeled source-city data and evaluated on
hidden benchmark data containing source-city and unseen target-city
scenes, establishing a privacy-preserving setting for cross-city
generalization.

The Training-as-a-Service setup distinguishes Track~6 from traditional
domain-adaptation benchmarks. Unsupervised domain-adaptation methods
require target-domain data during training
\cite{chen2018domain,li2022equalized,michaelis2019benchmarking}, an
assumption that does not hold in this setting. The problem therefore falls
within single-source domain generalization, where detectors must
generalize from labeled source data alone
\cite{ren2015faster,cui2019classbalanced,lin2021gan,
neubeck2006efficient,qi2024doubleaug}. Semi-supervised learning,
pseudo-label refinement, and augmentation can improve robustness
\cite{ruan2025finegrained,saito2019strong,li2020overcoming,
kisantal2019augmentation,sohn2020simple,tian2025yolov12,
wang2024yolov10}, while detector ensembles exploit complementary error
characteristics across architectures
\cite{danish2024improving,li2022crossdomain,milestone2025hafnia}.
These approaches have seldom been integrated into a unified framework for
privacy-preserving cross-city generalization.

Motivated by these challenges, this study proposes DRAFE, a
Domain-Robust Asymmetric Fusion Ensemble for cross-city fine-grained
traffic object detection. DRAFE combines complementary detection
transformers within a data-centric training and inference framework
organized around two premises. First, a scalable pseudo-label expansion
pipeline can increase annotated training coverage while maintaining
annotation quality through complete human review. Second, asymmetric
ensemble fusion with reliability-weighted coordinate fusion and
complementary hypothesis recovery can improve prediction quality under
cross-city domain shift.

The contributions of this work are as follows:

\begin{enumerate}
  \item DRAFE is proposed as a Domain-Robust Asymmetric Fusion Ensemble
        for cross-city fine-grained traffic object detection. The method
        combines asymmetric detector roles, class-consistent localization
        fusion, agreement-aware confidence recalibration, and
        complementary hypothesis recovery.

  \item A scalable pseudo-label expansion pipeline produces a corpus of
        6,049 images with 203,619 human-reviewed annotations. A
        meta-model trained on 1,683 manually annotated images proposes
        labels for corpus expansion, followed by complete human review,
        yielding 3.6 times greater image coverage.

  \item Evaluation on Track~6 of the 10th AI City Challenge shows that
        DRAFE achieves an mAP of 0.4022 and ranks sixth among 25
        participating teams. The method improves by 0.0553 mAP, or
        15.9\% relative, over a preliminary ensemble submission evaluated
        under identical benchmark conditions and by 0.0181 mAP, or 4.7\%
        relative, over the strongest standalone component detector.
\end{enumerate}

\section{Related Work}
\label{sec:related}

\subsection{Object Detection, Domain Generalization, and Data-Centric Learning}
\label{sec:related1}

Deep learning detectors have evolved from two-stage CNN architectures to
one-stage models and subsequently to transformer-based frameworks with
end-to-end set prediction and multi-scale feature representation
\cite{bochkovskiy2020yolov4,cai2018cascade,lin2017focal,
chen2024lwdetr,peng2025dfine,robinson2026rfdetr}.
Performance can degrade under domain shift from unseen environments.
Domain-adaptation methods align source and target distributions
\cite{chen2018domain,li2022crossdomain,michaelis2019benchmarking},
but require access to target-domain data during training. Single-source
domain generalization instead seeks representations that transfer from
labeled source data to unseen domains through feature diversification,
augmentation, and vision-language alignment
\cite{cui2019classbalanced,lin2021gan,neubeck2006efficient,
qi2024doubleaug,vidit2023clip}. Complementary data-centric strategies,
including pseudo-label refinement, semi-supervised learning, and
copy-paste augmentation, can extend training diversity and improve
robustness
\cite{ghiasi2021copypaste,gupta2019lvis,huang2025deim,
kisantal2019augmentation,sohn2020simple}.

Transportation studies have also demonstrated the value of systematic
model comparison and context-aware learning, including multi-model
evaluation for infrastructure-condition prediction
\cite{afriyie2025bridge}, spatially informed machine learning for
transportation-demand analysis \cite{terkper2025vehicle}, and
safety-aware reinforcement learning for adaptive traffic control under
changing work-zone conditions \cite{afriyie2026safety}.

\subsection{Detector Ensembles and Fine-Grained Vehicle Recognition}
\label{sec:related2}

Ensembling can improve localization and robustness by combining
complementary predictions from multiple models. Common fusion strategies
include non-maximum suppression, Soft-NMS, Non-Maximum Weighted
averaging, and Weighted Boxes Fusion. Weighted Boxes Fusion computes
confidence-weighted coordinate averages across overlapping detections
\cite{solovyev2021wbf}. Conventional Weighted Boxes Fusion clusters
spatially overlapping predictions without explicitly assigning anchor and
support roles. Although fixed model-level weights can be incorporated,
the method does not inherently impose anchor-conditioned one-to-one
matching, agreement-specific confidence recalibration, or explicit
recovery of unmatched support hypotheses.

Fine-grained vehicle recognition compounds these challenges. Najeeb et
al.~\cite{najeeb2022finegrained} examined confusion among visually
similar urban traffic classes, while Ruan et
al.~\cite{ruan2025finegrained} examined evidence fusion under degraded
conditions. The AI City Challenge series provides a benchmark for
intelligent transportation systems vision \cite{tang2025aicity9}.
Track~6 of the 10th edition addresses cross-city fine-grained object
detection through the Project Hafnia managed-training pipeline
\cite{aicity2026track6,milestone2024dataset,milestone2025hafnia,
milestone2026training}.

While Weighted Boxes Fusion \cite{solovyev2021wbf} and pseudo-label
generation \cite{sohn2020simple} are established techniques, DRAFE
introduces three departures from standard ensemble practice:

\begin{enumerate}
  \item Asymmetric anchor-support role assignment, in which one detector
        governs the initial hypothesis space.

  \item Anchor-conditioned class-consistent one-to-one matching, which
        prevents cross-class merging when detections from different
        semantic categories overlap spatially.

  \item Complementary hypothesis recovery, which retains selected
        unmatched support detections and thereby relaxes the
        anchor-recall ceiling.
\end{enumerate}

\section{Methodology}
\label{sec:method}

\subsection{Problem Formulation}
\label{sec:problem_formulation}

This study addresses fine-grained traffic-object detection under
geographic domain shift, as defined in AI City Challenge Track~6. A
detector is trained on source-domain data within the Hafnia workflow and
evaluated on hidden imagery containing source-city and target-city scenes
\cite{aicity2026track6}. The label space comprises ten categories: car,
pickup truck, single truck, combination truck, heavy-duty vehicle,
trailer, motorcycle, bicycle, van, and person.

The set of predictions returned by detector $m$ for image $x$ is
\begin{equation}
  D_m(x) = \left\{ \left(b_j,\,y_j,\,s_j\right) \right\}_{j=1}^{N_m},
  \label{eq:detector}
\end{equation}
where $b_j$ denotes a normalized axis-aligned bounding box, $y_j$
denotes the class label, $s_j$ denotes the confidence score, and $N_m$
is the number of detections returned by detector $m$.

Cross-city evaluation introduces two coupled challenges. Classification
and localization errors are not uniformly distributed across detectors,
and aggressive suppression can remove uncertain but correct predictions
in the target domain. DRAFE therefore preserves the hypothesis space of
the strongest high-recall model, exploits cross-model agreement for
localization refinement, and recovers complementary objects detected only
by support models.

\subsection{Overview of the DRAFE Framework}
\label{sec:overview}

DRAFE is structured as a training-to-fusion pipeline with three
independently trained detectors: two LW-DETR XLarge models and one
RF-DETR Base model. The two LW-DETR XLarge instances share the same
architecture but follow independent optimization trajectories, while
RF-DETR Base introduces architectural heterogeneity and a distinct
precision-recall profile. LW-DETR consists of a vision-transformer
encoder, feature projector, and shallow DETR decoder, with multilevel
feature aggregation and interleaved local-global attention
\cite{chen2024lwdetr}. RF-DETR is a real-time detection transformer
designed around an accuracy-latency tradeoff \cite{robinson2026rfdetr}.
Unlike symmetric fusion methods, DRAFE enforces anchor-conditioned
one-to-one matching, agreement-specific confidence recalibration, and
explicit recovery of unmatched support hypotheses.

All detectors undergo traffic-domain intermediate pretraining on a
curated corpus before challenge-specific fine-tuning within Hafnia.
After independent training, the strongest LW-DETR XLarge model is
assigned the anchor role, denoted \LWXA{}. The second LW-DETR XLarge
model, denoted \LWXB{}, provides within-family localization consensus.
RF-DETR Base, denoted \RFB{}, provides cross-architecture
complementarity. \Cref{tab:roles} summarizes these roles.

\begin{table}[htbp]
  \caption{Asymmetric role assignment of the constituent detectors.}
  \label{tab:roles}
  \centering
  \begin{tabular}{@{}lll@{}}
    \toprule
    Detector          & Role    & Primary contribution                       \\
    \midrule
    LW-DETR XLarge-A  & Anchor  & Dense, high-recall hypothesis set           \\
    LW-DETR XLarge-B  & Support & Localization consensus and additional recall \\
    RF-DETR Base      & Support & Architecturally complementary detections    \\
    \bottomrule
  \end{tabular}
\end{table}

\begin{figure}[htbp]
  \centering
  \includegraphics[width=\linewidth]{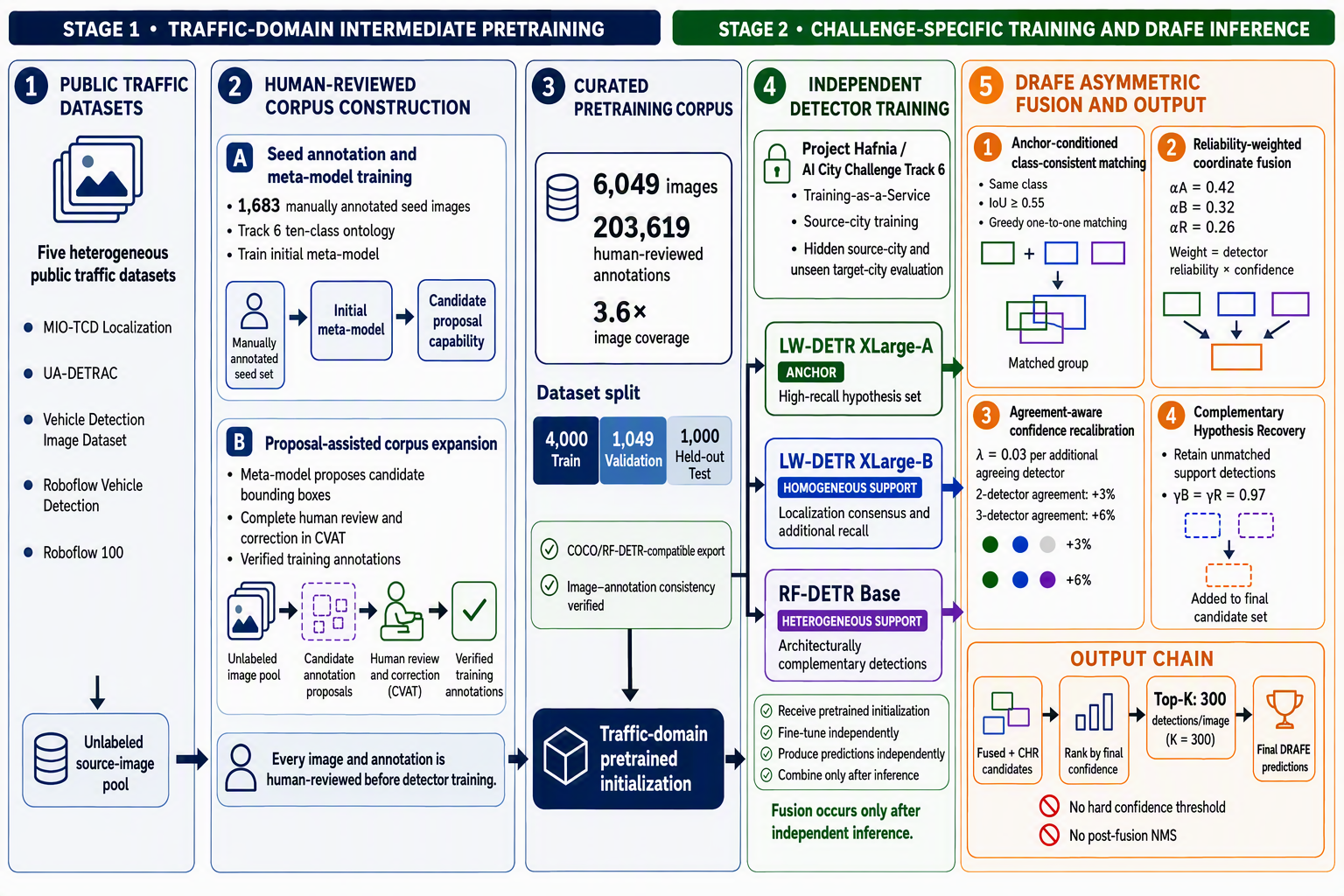}
  \caption{
    Overview of the DRAFE training and inference framework. During
    challenge-specific training, each Hafnia managed experiment combines
    the platform-hosted dataset with a trainer package, execution command,
    and selected GPU configuration.
  }
  \label{fig:pipeline}
\end{figure}

As illustrated in \cref{fig:pipeline}, \LWXA{} provides the anchor
hypotheses. \LWXB{} provides within-family localization consensus, while
\RFB{} contributes cross-architecture complementarity. The inference
procedure applies class-consistent anchor matching, reliability-weighted
coordinate fusion, agreement-aware confidence recalibration, and
complementary hypothesis recovery before retaining the 300
highest-ranked candidates.

\subsection{Traffic-Domain Intermediate Pretraining}
\label{sec:pretraining}

All component detectors undergo intermediate pretraining on a curated
traffic-domain corpus assembled from five public sources: MIO-TCD
Localization \cite{luo2018miotcd}, UA-DETRAC \cite{wen2020uadetrac},
the Vehicle Detection Image Dataset, the Roboflow vehicle-detection
dataset, and Roboflow~100 \cite{ciaglia2022roboflow100}. The corpus was
constructed through a two-stage annotation pipeline designed to increase
annotated coverage while controlling manual annotation effort.

\paragraph{Stage 1: Meta-model training on seed annotations.}
A representative set of 1,683 images was selected and fully annotated by
human reviewers using CVAT. Original source labels were not transferred
because the source taxonomies and bounding-box conventions differed. All
annotations followed the ten-class Track~6 ontology. Bicycle and
Motorcycle boxes were restricted to the vehicle body, with a separate
Person box assigned to each visible rider. Box pairs with
$\mathrm{IoU}\geq 0.90$ were automatically flagged for duplicate review.
The manually annotated seed set was then used to train a meta-model
whose sole role was to generate initial bounding-box proposals for a
larger image pool.

\paragraph{Stage 2: Pseudo-label expansion and human review.}
The meta-model generated candidate annotations for the 6,049-image
corpus. These predictions served only as initial annotation proposals.
Every image subsequently underwent complete human review and correction
in CVAT before inclusion in the training corpus. Therefore, no
automatically generated annotation was incorporated into detector
training without human verification.

The resulting corpus contains 203,619 verified object instances and
provides 3.6 times greater image coverage than the 1,683-image seed
set. The corpus was partitioned into 4,000 training images, 1,049
validation images, and 1,000 held-out test images. Car and Person
account for 84.5\% of all annotations, whereas Trailer and Heavy-Duty
Vehicle contain 390 and 391 instances, respectively. The reviewed
annotations were exported in COCO-compatible and RF-DETR-compatible
formats and checked for image-annotation consistency before intermediate
pretraining.

\subsection{Independent Detector Training}
\label{sec:training}

Intermediate pretraining on the curated public corpus was completed
outside the Hafnia challenge environment. Only pretrained model
initializations and challenge-specific training implementations were
incorporated into the trainer packages used for Hafnia fine-tuning. No
external images were introduced into the managed challenge environment.
Each detector was fine-tuned separately so that the component models
retained distinct optimization histories and prediction errors. The
ensemble was constructed from independently trained models rather than
through post hoc modification of a single detector.

At inference, each trained model independently processed every benchmark
image, and fusion was applied only after all three component detectors
completed inference.

\subsection{Anchor-Conditioned Class-Consistent Matching}
\label{sec:matching}

For each image, detections from \LWXA{} are sorted by descending
confidence and treated as anchor hypotheses. Let
$d_a=(b_a,y_a,s_a)$ denote an anchor detection with bounding box
$b_a$, class label $y_a$, and confidence $s_a$. The two support
detectors are indexed by $k\in\{B,R\}$, representing \LWXB{} and
\RFB{}, respectively. The $i$-th detection from support detector $k$
is $d_{k,i}=(b_{k,i},y_{k,i},s_{k,i})$.

Let $U_k$ denote the set of currently unused detections from detector
$k$. For each anchor $d_a$, the selected support match is
\begin{equation}
  d_k^{*}(a) = \underset{d_{k,i}\in U_k}{\arg\max}
    \;\mathrm{IoU}\left(b_a,\,b_{k,i}\right),
  \label{eq:matching}
\end{equation}
subject to
\begin{equation}
  y_{k,i}=y_a \quad\text{and}\quad
  \mathrm{IoU}\left(b_a,\,b_{k,i}\right)\geq \tau_f,
  \qquad \tau_f=0.55.
\end{equation}
If no eligible support detection exists, then
$d_k^{*}(a)=\varnothing$. Once assigned, a support detection is
removed from $U_k$, preventing the same support prediction from being
matched to multiple anchors. Class-consistent matching also prevents
cross-class merging when objects belonging to different categories
overlap spatially.

\subsection{Reliability-Weighted Coordinate Fusion}
\label{sec:fusion}

The matched group associated with anchor $d_a$ is
\begin{equation}
  G_a = \{d_a\} \cup
    \left\{ d_k^{*}(a) \mid k\in\{B,R\},\;d_k^{*}(a)\neq\varnothing \right\}.
\end{equation}
For each contributing detector $m$, the fusion weight is the product of
detector reliability $\alpha_m$ and detection confidence $s_m$:
\begin{equation}
  w_m=\alpha_m s_m.
\end{equation}
The selected reliability coefficients are $\alpha_A=0.42$,
$\alpha_B=0.32$, and $\alpha_R=0.26$. The fused bounding box is
\begin{equation}
  \hat{b}_a = \frac{
    \displaystyle\sum_{m\in G_a}w_m b_m
  }{
    \displaystyle\sum_{m\in G_a}w_m
  }.
  \label{eq:coordinate_fusion}
\end{equation}
Fusion is performed in corner-coordinate form $(x_1,y_1,x_2,y_2)$,
and the resulting coordinates are clipped to the normalized interval
$[0,1]$.

\subsection{Agreement-Aware Confidence Recalibration}
\label{sec:recalibration}

The fused confidence is designed for candidate ranking rather than
probabilistic interpretation. DRAFE retains the maximum confidence in
the matched group and applies a conservative multiplicative agreement
bonus:
\begin{equation}
  \hat{s}_a = \min\left\{
    1,\; s_{\max}\left[1+\lambda\left(|G_a|-1\right)\right]
  \right\},
  \qquad \lambda=0.03.
  \label{eq:recalibration}
\end{equation}
Agreement between two detectors produces a 3\% multiplicative increase,
while agreement among all three detectors produces a 6\% increase.

\subsection{Complementary Hypothesis Recovery}
\label{sec:chr}

Anchor-based fusion cannot recover an object that is absent from the
anchor hypothesis set. DRAFE therefore retains unused support detections
remaining in $U_k$ after matching, while conservatively scaling their
confidence:
\begin{equation}
  \hat{s}_{k,i} = \gamma_k s_{k,i},
  \qquad \gamma_B=\gamma_R=0.97,
  \qquad d_{k,i}\in U_k.
  \label{eq:chr}
\end{equation}
Because $\gamma_k<1$, anchor-supported hypotheses receive a modest
ranking advantage, while strong unmatched support detections remain
eligible for final selection.

\subsection{Budgeted High-Recall Selection}
\label{sec:budget}

The fused detections and complementary recovered hypotheses are pooled
and sorted by recalibrated confidence. The 300 highest-ranked detections
are retained:
\begin{equation}
  D_{\mathrm{final}}(x) = \operatorname{TopK}\!\left(
    D_{\mathrm{fused}}(x)\cup D_{\mathrm{CHR}}(x),\; K=300
  \right).
  \label{eq:topk}
\end{equation}
No hard confidence threshold is applied before final ranking, and no
additional non-maximum suppression is applied after fusion.

\subsection{Inference Algorithm and Configuration}
\label{sec:algorithm}

\Cref{alg:drafe} summarizes the DRAFE inference procedure.
\Cref{tab:config} reports the selected inference configuration.

\begin{algorithm}[htbp]
  \caption{Domain-Robust Asymmetric Fusion Ensemble}
  \label{alg:drafe}
  \begin{algorithmic}[1]
    \REQUIRE Predictions from \LWXA{}, \LWXB{}, and \RFB{} for image $x$
    \STATE Initialize unused support sets $U_B$ and $U_R$
    \STATE Initialize $D_{\mathrm{fused}}\gets\emptyset$
    \STATE Sort \LWXA{} detections by descending confidence
    \FOR{each anchor detection $d_a$}
      \STATE Initialize $G_a\gets\{d_a\}$
      \FOR{each support detector $k\in\{B,R\}$}
        \STATE Find the highest-IoU unused support detection satisfying
               the class constraint and $\mathrm{IoU}\geq0.55$
        \IF{an eligible support detection exists}
          \STATE Add the support detection to $G_a$
          \STATE Remove the support detection from $U_k$
        \ENDIF
      \ENDFOR
      \STATE Compute $\hat{b}_a$ using \cref{eq:coordinate_fusion}
      \STATE Compute $\hat{s}_a$ using \cref{eq:recalibration}
      \STATE Add $(\hat{b}_a,y_a,\hat{s}_a)$ to $D_{\mathrm{fused}}$
    \ENDFOR
    \STATE Scale unused support detections by $\gamma_k=0.97$
    \STATE Form $D_{\mathrm{CHR}}$ from the scaled support detections
    \STATE Pool and sort $D_{\mathrm{fused}}\cup D_{\mathrm{CHR}}$
    \STATE Return the 300 highest-ranked detections
    \ENSURE Final detection set $D_{\mathrm{final}}(x)$
  \end{algorithmic}
\end{algorithm}

\begin{table}[htbp]
  \caption{DRAFE inference configuration.}
  \label{tab:config}
  \centering
  \small
  \begin{tabular}{@{}ll@{}}
    \toprule
    Component                  & Setting                                     \\
    \midrule
    Anchor detector            & LW-DETR XLarge-A                            \\
    Support detectors          & LW-DETR XLarge-B; RF-DETR Base              \\
    Reliability weights        & $(0.42,\,0.32,\,0.26)$                      \\
    Class constraint           & Exact class identity                        \\
    Matching strategy          & Greedy one-to-one matching                  \\
    Fusion IoU threshold       & $0.55$                                      \\
    Agreement bonus            & $0.03$ per additional detector              \\
    Support confidence scale   & $0.97$                                      \\
    Post-fusion threshold      & None                                        \\
    Post-fusion NMS            & None                                        \\
    Final detection budget     & 300 detections per image                    \\
    \bottomrule
  \end{tabular}
\end{table}

\section{Experiments}
\label{sec:experiments}

\subsection{Dataset and Data Access}

Track~6 uses a traffic dataset provided through the Milestone Project
Hafnia platform
\cite{aicity2026track6,milestone2024dataset,milestone2025hafnia}. The
official training split contains 10,374 images, and the validation split
contains 2,564 images. The hidden benchmark contains 14,868 images from
source-city and unseen target-city scenes. The dataset contains
approximately 150,000 annotated instances across ten fine-grained traffic
categories.

The challenge data were hosted in Hafnia's Data Library and supplied to
managed training experiments. Participants could inspect dataset-level
information and sample data but could not download the underlying
challenge imagery. Track~6 required challenge training and benchmark
inference to be executed through managed experiments. Under the challenge
account configuration, only one experiment could run at a time, and
trained models and trainer packages were limited to 2~GB. Prediction
artifacts were retrieved from completed experiments and submitted to the
official AI City Challenge Evaluation Server.

\subsection{Evaluation Metrics}

The primary evaluation metric is mean Average Precision:
\begin{equation}
  \mathrm{mAP} = \frac{1}{C}\sum_{c=1}^{C}\mathrm{AP}_c,
  \label{eq:map}
\end{equation}
where $C$ is the number of object classes and $\mathrm{AP}_c$ is the
Average Precision for class $c$. The evaluation server additionally
reports $\mathrm{AP}_{50}$, $\mathrm{AP}_{75}$, scale-stratified AP,
and average recall at specified detection budgets.

\subsection{Implementation Details}

Each Hafnia experiment paired the platform-hosted challenge dataset with
a trainer package, an execution command, and a selected GPU
configuration. All challenge-specific experiments were conducted in the
Project Hafnia managed environment
\cite{milestone2025hafnia,milestone2026training}. The DRAFE ensemble
consists of two independently trained LW-DETR XLarge detectors
\cite{chen2024lwdetr}, denoted \LWXA{} and \LWXB{}, and one RF-DETR
Base detector \cite{robinson2026rfdetr}, denoted \RFB{}. Each model was
initialized from intermediate pretraining on the curated traffic corpus
before challenge-specific fine-tuning.

\paragraph{RF-DETR.}
RF-DETR Base was fine-tuned for two epochs using AdamW with a learning
rate of $1\times10^{-4}$, batch size 4, and gradient accumulation over
four steps, corresponding to an effective batch size of 16. Training
used the Hafnia Lite compute tier with one NVIDIA T4 GPU and 16~GB of
GPU memory.

\paragraph{LW-DETR.}
LW-DETR XLarge-A and XLarge-B were independently fine-tuned for five
and two epochs, respectively, using AdamW with a learning rate of
$5\times10^{-4}$ and batch size 1. The longer schedule produced the
stronger full-test result for XLarge-A, which was consequently assigned
the anchor role. XLarge-B retained an independent optimization history
and served as a homogeneous support detector.

Prediction files were written to the experiment artifact directory and
retrieved through Hafnia's experiment-output mechanism before submission
to the official evaluation server. The component detectors were executed
independently, and fusion was applied after completion of component
inference. Challenge jobs were executed sequentially because the account
configuration permitted one active experiment at a time. These
experiments demonstrate compatibility with the managed training
environment but do not establish real-time deployment performance.

\section{Experimental Results and Discussion}
\label{sec:results}

\subsection{Evaluation Protocol}
\label{sec:protocol}

Final model comparisons are based on scores returned by the official AI
City Challenge Evaluation Server for predictions generated on the
complete hidden benchmark of 14,868 images. Component-level
configuration studies used the Hafnia development split. Standalone
detector scores correspond to individual component submissions, while the
reported DRAFE result corresponds to the best ensemble submission. The
preliminary ensemble provides an earlier comparison configuration
evaluated through the same benchmark system.

\subsection{Standalone Detector and Baseline Performance}
\label{sec:standalone}

\LWXA{} achieved the strongest standalone full-test result among the
three component detectors, with 0.3841 mAP and 300 predictions per
image. \LWXB{} achieved 0.3822 mAP with 300 predictions per image,
while \RFB{} achieved 0.3836 mAP with an average of 77.66 predictions
per image. Although their aggregate mAP values were similar, \LWXB{}
and \RFB{} exhibited different recall and localization profiles. \LWXB{}
achieved higher $\mathrm{AR}_{100}$, while \RFB{} produced fewer
predictions per image. These differences motivated their use as
complementary support detectors.

\begin{table}[htbp]
  \centering
  \small
  \setlength{\tabcolsep}{3pt}
  \caption{Standalone and preliminary ensemble performance on the hidden test set.}
  \label{tab:standalone}
  \resizebox{\columnwidth}{!}{%
  \begin{tabular}{llccccccccccccc}
    \toprule
    Model & Role & mAP & AP$_{50}$ & AP$_{75}$ & AP$_S$ & AP$_M$ & AP$_L$
          & AR$@1$ & AR$@10$ & AR$_{100}$ & AR$_S$ & AR$_M$ & AR$_L$
          & Boxes/img \\
    \midrule
    Preliminary ensemble  & Comparison            & 0.3469 & 0.4732 & 0.3596
      & 0.0515 & 0.1864 & 0.4800 & 0.2843 & 0.4227 & 0.4368
      & 0.0840 & 0.2518 & 0.5779 & 300.00 \\
    LW-DETR XLarge-A      & Anchor                & 0.3841 & 0.5253 & 0.3998
      & 0.0557 & 0.2154 & 0.5276 & 0.4079 & 0.6467 & 0.6881
      & 0.2323 & 0.4892 & 0.8193 & 300.00 \\
    LW-DETR XLarge-B      & Homogeneous support   & 0.3822 & 0.5210 & 0.4051
      & 0.0452 & 0.2015 & 0.5443 & 0.3957 & 0.6349 & 0.6709
      & 0.2195 & 0.4571 & 0.8143 & 300.00 \\
    RF-DETR Base          & Heterogeneous support & 0.3836 & 0.5254 & 0.4048
      & 0.0476 & 0.2070 & 0.5232 & 0.3868 & 0.5916 & 0.6169
      & 0.1143 & 0.3820 & 0.7603 & 77.66  \\
    \bottomrule
  \end{tabular}}
\end{table}

\subsection{Effect of Asymmetric Ensembling}
\label{sec:hetero}

DRAFE achieved 0.4022 mAP on the complete hidden benchmark. This
represents an absolute improvement of 0.0553 mAP over the preliminary
ensemble and 0.0181 mAP over the strongest standalone component,
\LWXA{}. Improvements were observed across the reported object-size
strata and at $\mathrm{AR}_{100}$.

\begin{table}[htbp]
  \centering
  \small
  \setlength{\tabcolsep}{4pt}
  \caption{
    Full-test performance of the preliminary ensemble, strongest
    standalone component, and DRAFE.
  }
  \label{tab:hetero}
  \resizebox{\columnwidth}{!}{%
  \begin{tabular}{lccccccc}
    \toprule
    Method & mAP & AP$_{50}$ & AP$_{75}$ & AP$_S$ & AP$_M$ & AP$_L$
           & AR$_{100}$ \\
    \midrule
    Preliminary ensemble           & 0.3469 & 0.4732 & 0.3596 & 0.0515
      & 0.1864 & 0.4800 & 0.4368 \\
    \LWXA{}                        & 0.3841 & 0.5253 & 0.3998 & 0.0557
      & 0.2154 & 0.5276 & 0.6881 \\
    \textbf{DRAFE}                 & \textbf{0.4022} & \textbf{0.5333}
      & \textbf{0.4259} & \textbf{0.0571} & \textbf{0.2335}
      & \textbf{0.5460} & \textbf{0.6981} \\
    \midrule
    Difference from preliminary ensemble & +0.0553 & +0.0601 & +0.0663
      & +0.0056 & +0.0471 & +0.0660 & +0.2613 \\
    Difference from \LWXA{}              & +0.0181 & +0.0080 & +0.0261
      & +0.0014 & +0.0181 & +0.0184 & +0.0100 \\
    \bottomrule
  \end{tabular}}
\end{table}

The gain over \LWXA{} was largest at $\mathrm{AP}_{75}$, where DRAFE
improved by 0.0261. This result is consistent with improved localization
among matched detections, although the aggregate benchmark metrics do not
independently isolate the contribution of each fusion operation.

\subsubsection{Qualitative Comparison}
\label{sec:qualitative}

\begin{figure}[htbp]
  \centering
  \includegraphics[width=\textwidth]{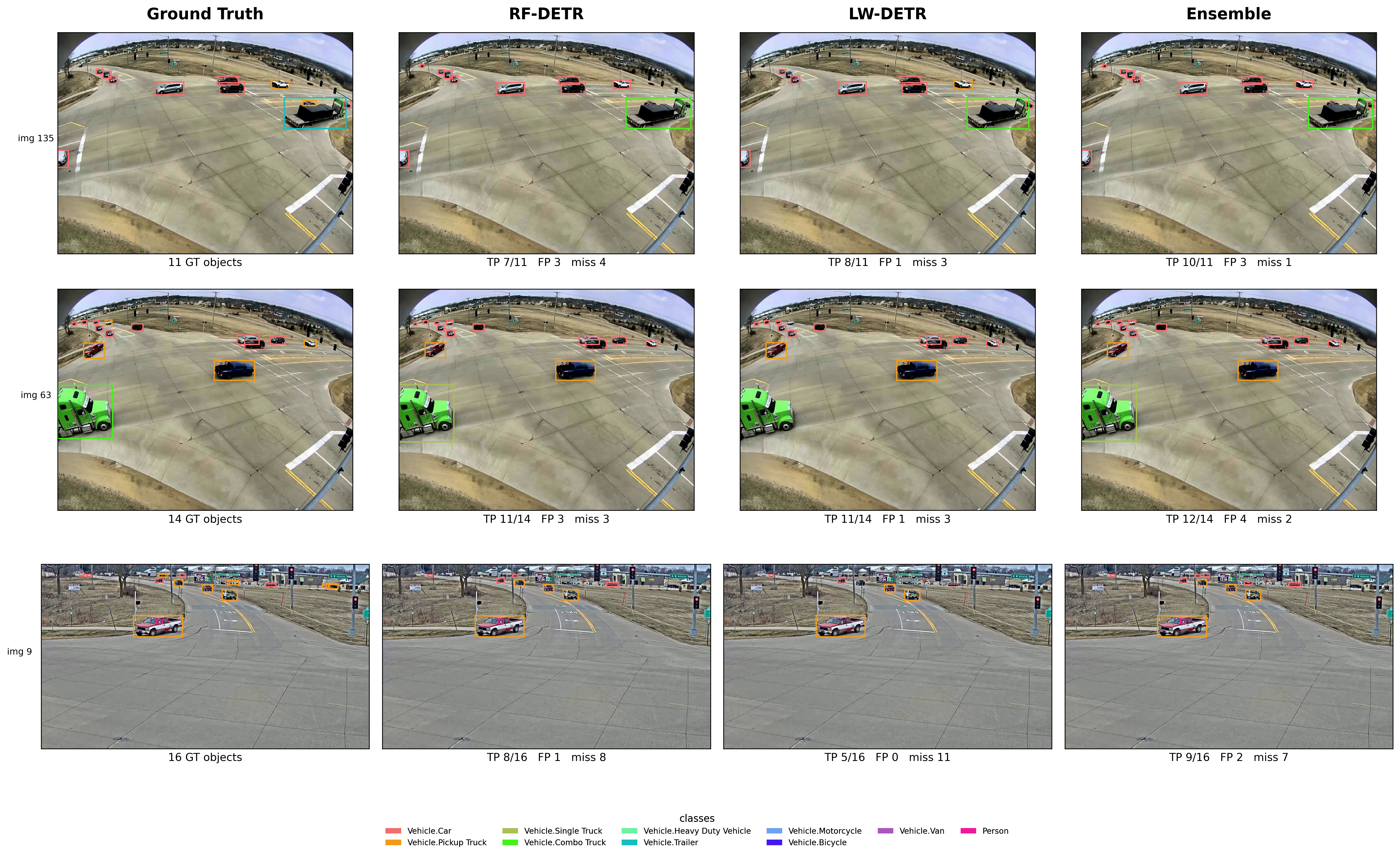}
  \caption{Qualitative comparison of standalone detectors and DRAFE.}
  \label{fig:ensemble-comparison}
\end{figure}

\Cref{fig:ensemble-comparison} compares ground-truth annotations with
predictions from RF-DETR, LW-DETR, and DRAFE on three selected images.
Across these examples, DRAFE recovers more ground-truth objects than
either standalone detector. The ensemble increases the true-positive
count from 7--8 to 10 in the first image, from 11 to 12 in the second
image, and from 5--8 to 9 in the third image. The recovered objects are
accompanied by additional false-positive hypotheses, reflecting the
precision-recall tradeoff associated with the high-recall
candidate-retention strategy.

\subsection{Ablation Studies}
\label{sec:ablation}

The ablation comparisons were conducted on the Hafnia development split
during the challenge period and should not be directly compared with the
full-test results reported above.

\paragraph{Model composition.}
Adding a fourth detector as a fusion-only support model did not improve
overall development performance. The four-model variant increased
$\mathrm{AP}_{75}$ from 0.4312 to 0.4321 and $\mathrm{AR}_{100}$ from
0.6968 to 0.6999, but reduced mAP from 0.4074 to 0.4068. The experiment
used a single training seed, and multi-seed evaluation was not conducted
within the challenge resource allocation.

\begin{table}[htbp]
  \centering
  \small
  \caption{Development-set ablation on ensemble size.}
  \label{tab:composition}
  \resizebox{\columnwidth}{!}{%
  \begin{tabular}{lccccccc}
    \toprule
    Configuration              & mAP    & AP$_{50}$ & AP$_{75}$ & AP$_S$
                               & AP$_M$ & AP$_L$    & AR$_{100}$ \\
    \midrule
    Three-model DRAFE          & \textbf{0.4074} & \textbf{0.5398} & 0.4312
      & \textbf{0.0604} & \textbf{0.2312} & \textbf{0.5542} & 0.6968 \\
    Four-model support ensemble & 0.4068 & 0.5379 & \textbf{0.4321}
      & 0.0584 & 0.2245 & 0.5528 & \textbf{0.6999} \\
    \bottomrule
  \end{tabular}}
\end{table}

\paragraph{Agreement bonus and reliability weights.}
Increasing the agreement bonus from $\lambda=0.03$ to $\lambda=0.04$
increased $\mathrm{AP}_{50}$ and $\mathrm{AR}_{100}$ but reduced
$\mathrm{AP}_{75}$, medium-object AP, large-object AP, and overall mAP.
Shifting one percentage point of reliability weight from \LWXA{} to
\RFB{}, from $(0.43,\,0.32,\,0.25)$ to $(0.42,\,0.32,\,0.26)$,
increased the development score to 0.4077.

\paragraph{Detection budget.}
A thresholding and suppression configuration that reduced output to
23.22 detections per image produced a development mAP of 0.3520. The
final configuration consequently retained its high-recall budget of 300
ranked detections per image without a hard preselection threshold.

\begin{table}[htbp]
  \centering
  \small
  \caption{
    AI City Challenge 2026 Track~6 leaderboard based on each team's best
    reported full-test score.
  }
  \label{tab:leaderboard}
  \resizebox{\columnwidth}{!}{%
  \begin{tabular}{clcccccc}
    \toprule
    Rank & Team                             & mAP    & AP$_{50}$ & AP$_{75}$
         & AP$_S$ & AP$_M$ & AP$_L$ \\
    \midrule
    1  & SKKU-AL-T1                       & 0.4753 & 0.6207 & 0.5031
       & 0.0920 & 0.3041 & 0.6110 \\
    2  & BIT-ODL                          & 0.4281 & 0.5539 & 0.4568
       & 0.0637 & 0.2639 & 0.5669 \\
    3  & Remote Vibecoders from Chisinau  & 0.4176 & 0.5727 & 0.4521
       & 0.0790 & 0.2739 & 0.5339 \\
    4  & BK2TheFuture                     & 0.4169 & 0.5289 & 0.4407
       & 0.0591 & 0.2210 & 0.5548 \\
    5  & S2 Detection                     & 0.4114 & 0.5447 & 0.4414
       & 0.0747 & 0.2596 & 0.5407 \\
    \midrule
    6  & \textbf{VisionOps/DRAFE}         & \textbf{0.4022}
       & \textbf{0.5333} & \textbf{0.4259} & \textbf{0.0571}
       & \textbf{0.2335} & \textbf{0.5460} \\
    \midrule
    7  & Team IPCV                        & 0.3882 & 0.5021 & 0.4068
       & 0.0578 & 0.1735 & 0.5307 \\
    8  & bfc                              & 0.3741 & 0.5097 & 0.3989
       & 0.0445 & 0.2369 & 0.5048 \\
    9  & NextITS                          & 0.3654 & 0.4879 & 0.3857
       & 0.0660 & 0.1924 & 0.5034 \\
    10 & Team United                      & 0.3529 & 0.4655 & 0.3656
       & 0.0487 & 0.1588 & 0.4675 \\
    \bottomrule
  \end{tabular}}
\end{table}

\subsection{Object-Scale Analysis}
\label{sec:scale}

Performance varied substantially by object size. Large objects achieved
$\mathrm{AP}_L=0.5460$ and $\mathrm{AR}_L=0.8305$, while medium objects
achieved $\mathrm{AP}_M=0.2335$ and $\mathrm{AR}_M=0.4938$. Small
objects achieved $\mathrm{AP}_S=0.0571$ and $\mathrm{AR}_S=0.2131$.
The benchmark reports aggregate scale-stratified metrics without
per-class or error-type decomposition. Consequently, the available
results establish that small-object performance is substantially lower
than medium- and large-object performance but do not identify the
specific class, localization, or confusion mechanisms responsible for
that difference.

\section{Conclusion}
\label{sec:conclusion}

This study introduced DRAFE, a Domain-Robust Asymmetric Fusion Ensemble
for cross-city fine-grained traffic-object detection under
privacy-preserving Training-as-a-Service constraints. The method combines
traffic-domain intermediate pretraining with anchor-conditioned
class-consistent matching, reliability-weighted coordinate fusion,
agreement-aware confidence recalibration, and complementary hypothesis
recovery. The intermediate-pretraining pipeline produced a 6,049-image
corpus containing 203,619 human-reviewed annotations. On the complete
Track~6 hidden benchmark, DRAFE achieved 0.4022 mAP and ranked sixth
among 25 participating teams. The ensemble improved by 0.0181 mAP over
the strongest standalone component detector and by 0.0553 mAP over a
preliminary ensemble evaluated through the same benchmark system.

The results provide evidence that detector complementarity and asymmetric
inference refinement can improve cross-city fine-grained object detection
in a managed training environment. Further work should examine
small-object representation, class-specific error patterns, calibration
across geographic domains, and multi-seed evaluation of the component
detectors and fusion configuration.

\bibliographystyle{splncs04}
\bibliography{main}

\end{document}